\documentclass{edm_article}

\begin{document}
\emergencystretch=3em 

\title{Comparing Human Role-Players and LLM-Simulated Clients in Online Counselling Training: An Analysis of Counselling Patterns}

\numberofauthors{3}
\author{
Eric Rudolph, Philipp Steigerwald, Jens Albrecht\\
\affaddr{Technische Hochschule Nürnberg Georg Simon Ohm}\\
\affaddr{Nürnberg, Germany}\\
\email{\{eric.rudolph, philipp.steigerwald, jens.albrecht\}@th-nuernberg.de}
}

\maketitle

\begin{abstract}
This study investigates the capabilities of Large Language Models to simulate counselling clients in educational role-plays in comparison to human role-players.
Initially, we recorded role-playing sessions, where novice counsellors interacted with human peers acting as clients, followed by role-plays between humans and clients simulated by Mistrals Mixtral 8x7b using 4-bit quantization. These interactions were analysed with a counselling communication pattern system at sentence level. We investigated two key questions: (1) to what extent LLM-generated responses replicate authentic conversational dynamics and (2) whether counsellors’ communication behaviour differs when interacting with human versus LLM-simulated clients. The findings highlight both similarities and differences in the application of counselling patterns across scenarios, showing the potential of LLM-based role-playing exercises to enhance counselling competencies and to identify areas for further refinement in virtual client simulations.

\end{abstract}

\keywords{Large Language Models, Counselling, Natural Language Processing, Educational Data Mining, AI in Education, Counselling Training, Client Simulation} 

\section{Introduction}
Text-based online counselling has gained increasing prominence as a flexible and accessible alternative to traditional face-to-face therapy, especially in reaching diverse populations with unique needs \cite{dwyer_suitability_2021,moock_support_2014}.  
Within counsellor education, role-playing exercises are central to developing crucial communication competencies \cite{osborn_role-playing_2013, nair_role_2019}, offering trainees controlled, practice-oriented scenarios to explore, reflect on, and refine their skills \cite{ronning_use_2019,shurts_preparing_2006}.  
However, role-plays with human peers often face logistical constraints such as coordinating schedules and achieving consistent complexity across multiple sessions \cite{yee_challenges_2024}, while also being limited by the role-play partner's availability, engagement level and ability to maintain character depth and consistency throughout the interactions \cite{maurya_using_2024}.

With the ongoing advancements in large language models (LLMs), the prospect of simulating client roles using AI-driven conversation agents has become an attractive avenue \cite{li_leveraging_2024,kerr_realist_2021,louie_roleplay-doh_2024}.  
In principle, these virtual clients offer standardization, flexibility, and round-the-clock availability, allowing trainees to practice and experiment in a safe, repeatable environment \cite{baker_virtual_2023, shorey_virtual_2019}.  
Despite the practical advantages, key questions remain regarding the authenticity of the learning experience \cite{stamer_artificial_2023}.  
Do LLM-generated clients exhibit realistic conversational dynamics, and do they elicit the same communication patterns from counsellors as human interlocutors would?

This paper examines text-based role-play counselling sessions, comparing interactions between human counsellors and human clients with those between human counsellors and LLM-simulated clients.
We draw on a specialised categorisation framework for counselling communications, analy\-sing the distribution and temporal evolution of specific interaction patterns in both settings.  
Our findings highlight significant similarities and differences in how clients disclose their concerns and how counsellors respond and structure the session.  
In doing so, this study provides insights into the potential of LLM-based role-plays to augment --- and possibly transform --- the way future counsellors are trained, while also identifying areas where further development is needed to narrow the gap between virtual and human clients.

\section{Related Work}

The integration of LLMs into counselling education and training has gained increasing attention.
In particular, AI-driven role-playing simulations are being explored as a means to enhance the learning experience of trainee counsellors by providing standardised, risk-free, and adaptive training environments.

A recent study examined the impact of role-play interactions between humans and an LLM-powered agent on educational outcomes \cite{chen_investigating_2025}.
This research compared Human-LLM role-play interactions with traditional Human-Human role-play settings, assessing their effectiveness in improving participants’ communication skills.
The findings indicated that while LLMs can facilitate structured and engaging role-play scenarios, there are still differences in spontaneity and contextual adaptation when compared to human interlocutors. A study leveraged GPT-4 to generate and analyse psychological counselling sessions through role-play interactions between human counsellors and LLM-simulated clients \cite{inaba_can_2024}.
Professional counsellors evaluated the AI-generated responses against human client responses in identical counselling contexts.
The findings demonstrated GPT-4's capability to effectively replicate conversational patterns, suggesting potential for LLM-based training simulations, though challenges in response consistency and emotional authenticity indicated the need for further refinements. VirCo (Virtual Client for Online Counselling) demonstrates another practical application of LLMs in counsellor training \cite{rudolph_ai-based_2025, rudolph_automated_2024}.
The system leverages an open-source LLM with persona descriptions and role-play transcripts, showing high coherence in evaluation while addressing privacy concerns through its architecture. A systematic review investigated the broader applications of LLMs in mental health education, particularly in psychotherapy training \cite{bhatt_digital_2024}. 
The analysis revealed that AI-driven simulations enable effective practice of communication strategies and provide immediate feedback in risk-free environments, though success depends on proper integration with established educational frameworks. 
Similarly, another study explored the opportunities and risks associated with deploying LLMs in counselling education \cite{lawrence_opportunities_2024}.
While LLMs have the potential to provide consistent and data-driven training experiences, concerns regarding bias, ethical considerations, and misinformation remain. 
These challenges are particularly critical in educational contexts where reliable learning outcomes must be ensured.

To the best of our knowledge, this is the first study to directly compare human–human and human–LLM counselling conversations using a standardized category system like GeCCo. Prior works have either demonstrated the feasibility of LLM-simulated clients or discussed their general benefits and challenges, but none have quantitatively contrasted the interaction patterns of AI-driven clients with those of real humans in a training context. Our study expands upon the existing literature by bridging that gap: We integrate the advances of an LLM-based virtual client platform (based on \cite{rudolph_ai-based_2025}) with a fine-grained communication analysis methodology (GeCCo) to evaluate how AI-simulated dialogues differ from and resemble traditional role-plays. 

\section{Dataset}

The dataset consists of two types of educational role-plays: Human-Human and Human-LLM interactions. The \textbf{Human-Human role-plays} were conducted as part of a degree program in social work and focused on online counselling fundamentals. A total of 64 chat-based counselling sessions were conducted via a chat platform, with one student acting as the client. Students selected a case study to shape their role. The \textbf{Human-LLM role-plays} were recorded in a learning platform to practice chat counselling skills.  The learning platform is based on the methodology described in VirCo \cite{rudolph_ai-based_2025}. While the prompt stayed the same as described in VirCo the language model has been exchanged from Vicuna 13B to Mixtral 8x7B.
Two distinct student groups participated: one from the online counselling course and another comprising students from a different program at a partnering university. As LLM Mistrals Mixtral 8x7B with 4-bit quantization has been chosen because its efficient inference enabled by the mixture-of-experts architecture, ensures fast responses. This is especially important when having multiple requests at the same time. We also selected a single model to maintain comparability between the Human-Human and Human-LLM datasets, ensuring similar sentence counts and minimizing confounds in our analysis.
In total, the data set contains 74 completed Human-LLM counselling conversations. 
We used the same case studies as in the Human-Human role plays by integrating them into the system prompt of the language model.

\begin{table}[ht]
    \centering
    \renewcommand{\arraystretch}{1.3}
    \caption{Comparison of Human-Human and Human-LLM Counselling Messages}
    \begin{tabular}{lcc}
        \hline
        \textbf{Statistic} & \textbf{Human-Human} & \textbf{Human-LLM} \\
        \hline
        Number of Conversations & 64 & 74 \\
        Number of Messages ($N$) & 2411 & 2892 \\
        Total Sentences & 5383 & 7040 \\
        Avg. Sentences per Msg & 2.23 & 2.43 \\
        Total Words & 59805 & 82587 \\
        Avg. Words per Msg & 24.81 & 28.56 \\
        \hline
    \end{tabular}
    
    \label{tab:human_vs_llm}
\end{table}

\section{Method}
Central to our analytical approach is the GeCCo (German E-Counselling Conversation Dataset) category system \cite{albrecht_gecco_2024}, a specialised framework designed for analysing online counselling conversations. Structured hierarchically across up to five layers of abstraction, GeCCo enables a highly granular classification of both counsellor and client communications. In its fifth layer, the system details 40 distinct counsellor categories and 28 client categories, capturing both broad dimensions (such as “Formalities” or “Impact factors”) and subtle communicative nuances.

GeCCo was selected because it synthesises several established categorisation frameworks, including Motivational Interviewing, and integrates insights from diverse counselling models. Given that our dataset comprises German-language counselling conversations—and considering that GeCCo is based on a German dataset—the system aligns well with our needs. Moreover, the category system has been successfully trained using a GBERT Large classifier, which was further fine-tuned with synthetic data from GPT-4 to classify our conversations effectively. The analysis was then conducted on layer 3 of the GeCCo hierachy (layer~1: most abstract; layer~5: finest granularity). The following paragraph shows an explanation of the used categories:

\noindent\textbf{Client Categories:} \\
\textbf{Analysis \& Agreement on Counselling Goals:} Statements involving the formulation and mutual agreement on counselling goals. \\
\textbf{Analysis \& Clarification of Problems:} Expressions aimed at exploring and clarifying the client’s issues. \\
\textbf{Help, Problem-Solving:} Utterances intended to facilitate problem resolution or offer direct assistance. \\
\textbf{Generating Motivation:} Strategies that evoke or enhance the client’s intrinsic motivation for change. \\
\textbf{Resource Activation:} Identification and mobilization of internal or external resources to support the client. \\
\textbf{Other Statements:} Miscellaneous expressions that do not fit into the main categories. \\
\textbf{Formalities at the Beginning:} Opening formalities such as greetings and introductory remarks. \\
\textbf{Formalities at the End:} Closing formalities including farewells and session summaries. \\[1ex]
\noindent\textbf{Counselor Categories:} \\
\textbf{Other Statements:} Counselor’s miscellaneous expressions that do not fit into the main categories. \\
\textbf{Conversation Opening:} Counselor-led initiations that set the tone and structure for the session. \\
\textbf{Moderation:} Efforts by the counselor to steer and structure the conversation effectively. \\
\textbf{Formalities at the Beginning:} Counselor's introductory remarks and greetings to initiate the session. \\
\textbf{Formalities at the End:} Counselor's closing remarks and farewells to conclude the session. \\
\textbf{Analysis \& Agreement on Counselling Goals:} Counselor’s statements involving the formulation and mutual agreement on counselling goals. \\
\textbf{Analysis/ Clarification-Reflection (Emotion):} Reflective responses aimed at clarifying the emotional content of the dialogue. \\
\textbf{Analysis/ Clarification-Reflection (Factual):} Reflective responses focused on clarifying factual details shared during the session. \\
\textbf{Help, Problem-Solving:} Expressions intended to assist in problem resolution or to provide direct help. \\
\textbf{Generating Motivation:} Strategies to evoke or enhance the client’s intrinsic motivation. \\
\textbf{Resource Activation:} Identification and mobilization of resources to support the client.

The analysis shows an examination of pattern distributions by computing the frequency of each GeCCo category in both Human-Human and Human-LLM interactions. This calculation identifies differences in the manifestation of counselling goals, problem clarification, and other communicative patterns. A temporal analysis segments conversations into deciles, revealing how interaction patterns—such as factual reflection and problem solving—vary across different stages of the counselling session.

A semantic analysis employs embedding-based representations, where embeddings for each message are generated using the GBERT classifier and visualised via t-distributed Stochastic Neighbour Embedding (t-SNE) \cite{maaten_visualizing_2008}. The resulting projections uncover clusters corresponding to various categories, such as “Analysis \& Clarification of Problems” and “Formalities,” thereby elucidating the semantic structure of both human and LLM-generated responses.

\section{Results}
This section presents the findings of a comparative analysis of Human-Human and Human-LLM counselling role-plays. It examines differences in communication patterns using three complementary approaches: (1) Pattern Distribution Analysis, which quantifies the frequency of different counselling behaviours across both interaction types; (2) Temporal Analysis, which investigates how interaction patterns evolve over the course of a conversation; and (3) Semantic Analysis, which visualises conversational similarities and differences using embedding-based representations.
\subsection{Pattern Distribution Analysis}
\begin{figure}
    \centering
    \includegraphics[width=1\linewidth]{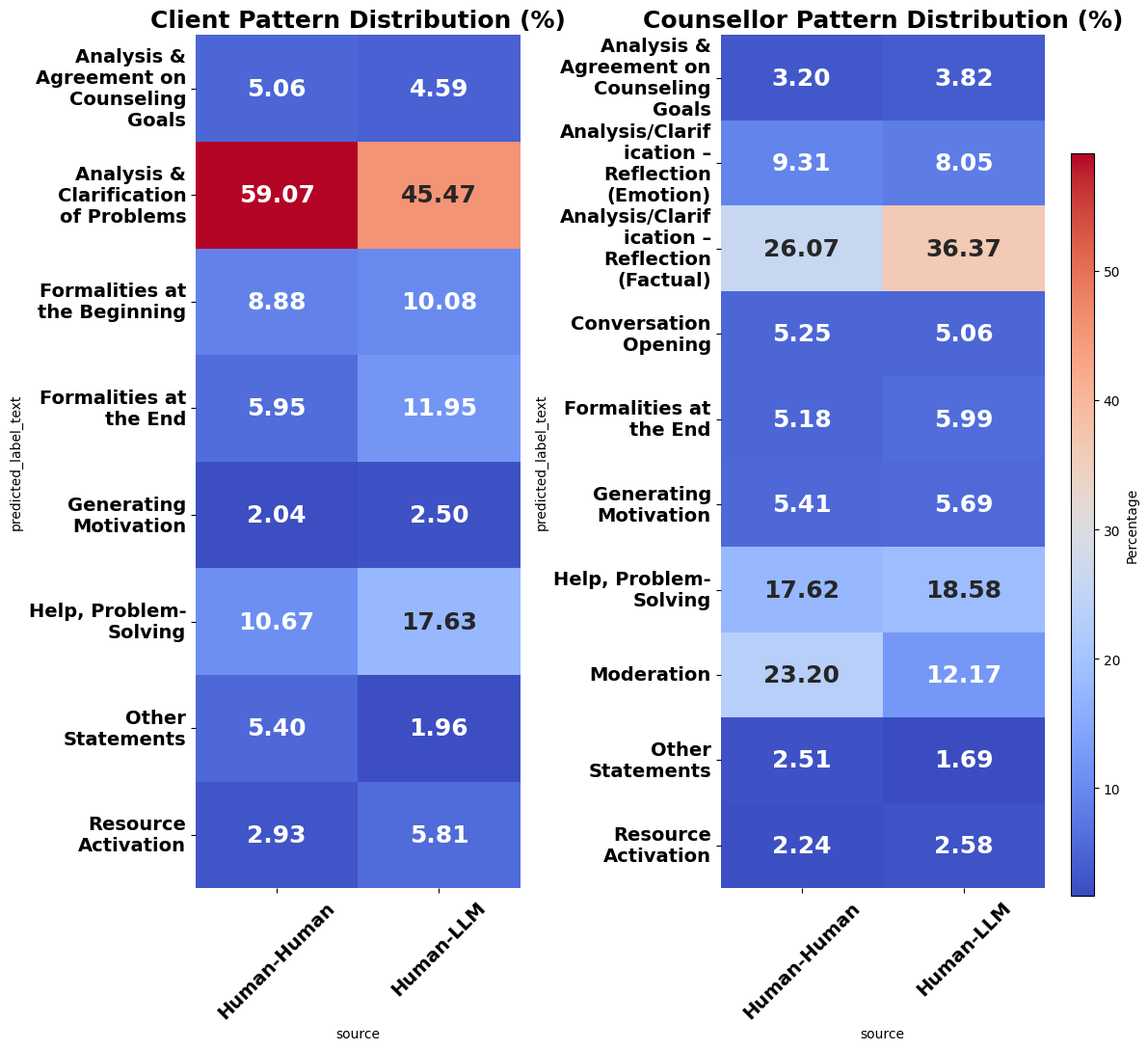}
    \caption{Pattern Distribution for Client and Counselor Patterns across all conversations}
    \label{fig:pattern-distribution}
    \Description{A grouped bar chart showing, for both clients and counsellors, the percentage use of each communication pattern in Human–Human versus Human–LLM role-plays. Client bars show that human clients use Analysis \& Clarification of Problems more often, while LLM clients use Help/Problem-Solving and Formalities at the End more. Counsellor bars show that human counsellors engage more in Moderation, whereas counsellors with LLM clients use Reflection (Factual) more.}
\end{figure}

The distribution of the client pattern shows a significant difference between Human-Human and Human-LLM interactions in the category \textbf{Analysis \& Clarification of problems} (Figure~\ref{fig:pattern-distribution}). Human clients use this pattern significantly more often (59.07\%) than LLM-simulated clients (45.47\%). This discrepancy results from the fact that LLM-simulated clients tend to disclose the problems embedded in the prompt more readily, which enables a faster transition to problem solving.

In addition, category \textbf{Help, Problem-Solving} occurs more frequently with LLM-simulated clients (17.63\%) than with human clients (10.67\%), suggesting that simulated clients take a more solution-focused approach. Similarly, \textbf{Formalities at the End} are more common in LLM-generated clients (11.95\%) than in human clients (5.95\%), probably reflecting structured conversation patterns typical of LLM responses. The counsellor distribution shows significant differences in the category \textbf{Analysis/Clarification--Reflection (Factual)}, which is more pronounced in Human-LLM interactions (36.37\%) compared to Human-Human conversations (26.07\%). This suggests that counsellors interacting with LLM clients focus more on factual reasoning, which may compensate for the LLM's tendency to disclose rather than explain problems. On the other hand, \textbf{moderation}, which includes facilitation and structuring responses, is significantly higher in Human-Human interactions (23.20\%) compared to Human-LLM conversations (12.17\%). This may suggest that counsellors in Human-Human interactions take a more active role in leading the discussion, whereas those conversing with LLM clients rely more on reflective techniques.

Other notable differences include \textbf{Help, Problem-Solving}, which remains relatively stable across conditions (17.62\% in Human-Human interactions vs. 18.58\% in Human-LLM), and \textbf{Conversation Opening}, which occurs slightly more frequently in Human-Human interactions.

\subsection{Temporal Analysis}
\begin{figure}
    \centering
    \includegraphics[width=1\linewidth]{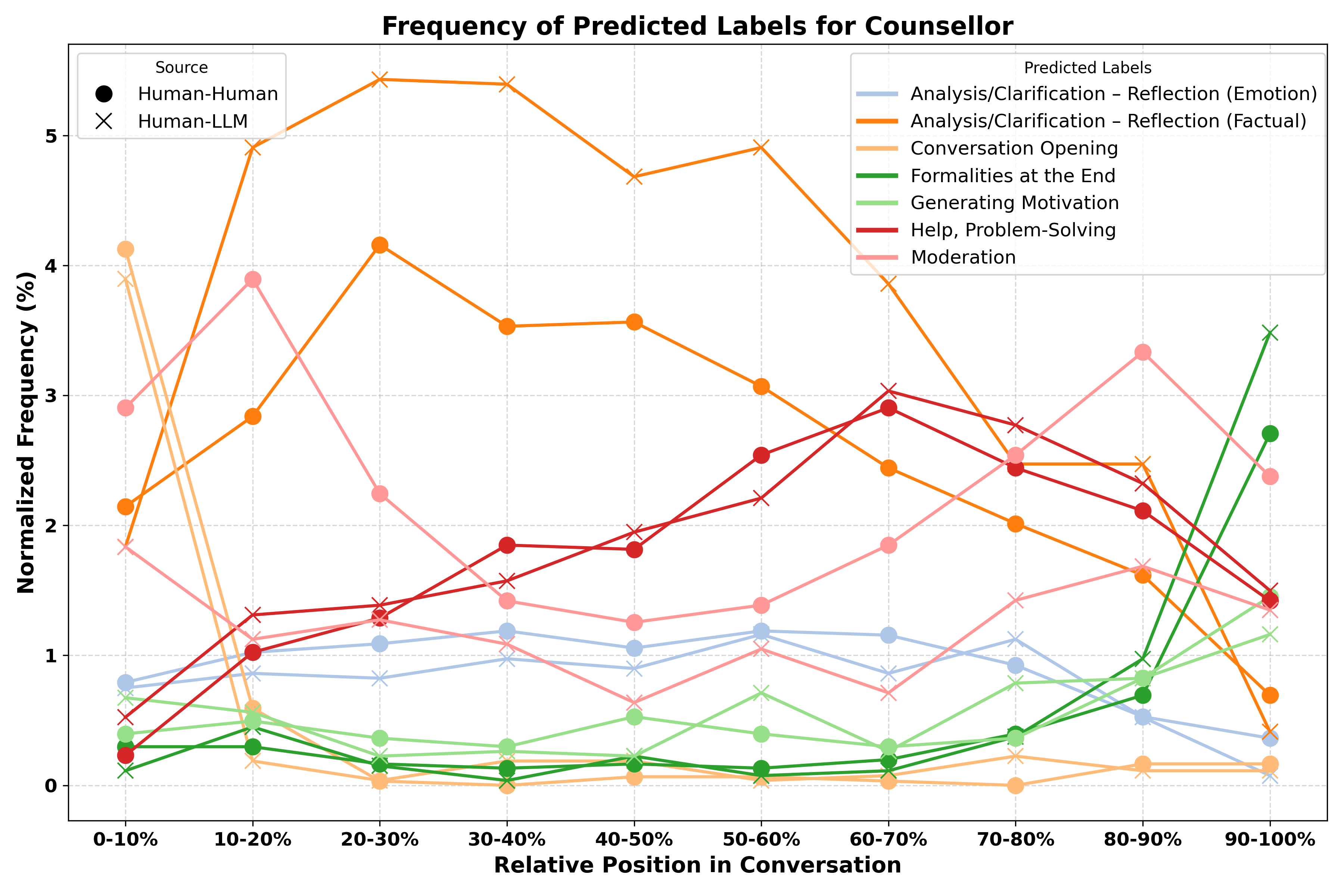}
    \caption{Frequency of Predicted Counsellor Labels along Conversation Progression}
    \label{fig:counsellor-temporal}
    \Description{A multi-line chart plotting the normalized frequency of counsellor communication patterns across ten equally-sized segments (deciles) of each session, for Human–Human and Human–LLM interactions. Lines represent patterns such as Analysis/Clarification–Reflection (Factual), Help/Problem-Solving, Moderation, Generating Motivation, Conversation Opening, and Formalities at Beginning/End, illustrating how each pattern rises or falls over the course of the session.}

\end{figure}

\begin{figure}
    \centering
    \includegraphics[width=1\linewidth]{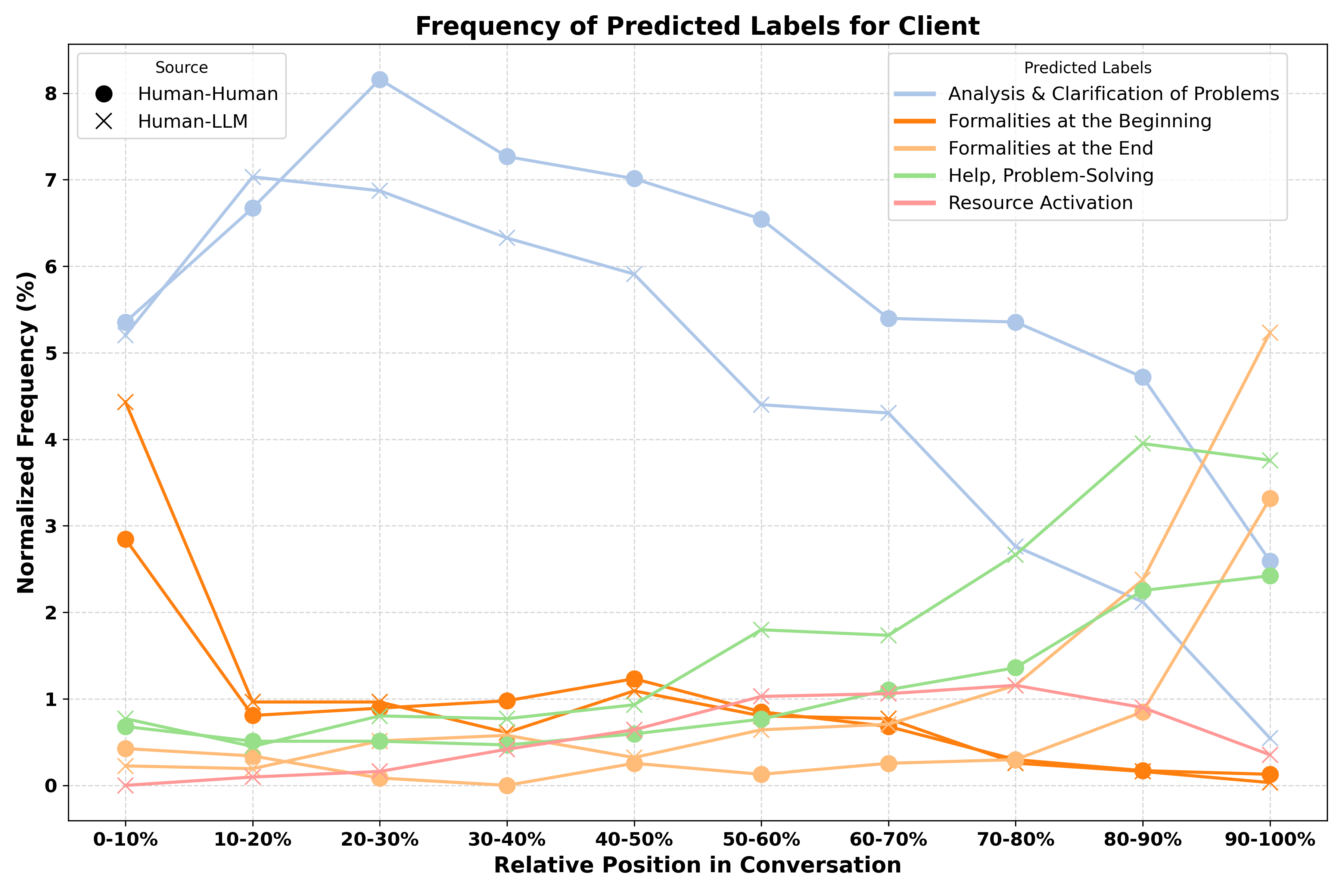}
    \caption{Frequency of Predicted Client Labels along conversation progression}
    \label{fig:client-temporal}
    \Description{A multi-line chart showing the normalized frequency of client communication patterns over ten deciles of session progression for both Human–Human and Human–LLM conversations. Lines denote Analysis & Clarification of Problems, Help/Problem-Solving, Resource Activation, Generating Motivation, Formalities at Beginning/End, and Other Statements, highlighting differences in how clients disclose issues and shift toward solutions over time.}
\end{figure}

Figures \ref{fig:counsellor-temporal} and \ref{fig:client-temporal} illustrate the temporal evolution of the predicted interaction labels for counsellors and clients across the relative position in conversations. The x-axis represents the conversation progression in deciles (0-10\% to 90-100\%), while the y-axis shows the normalised frequency (\%) of each pattern. These graphs can be used to examine how the interaction patterns evolve over time and how they differ between Human-Human and Human-LLM counselling sessions.

In the counsellors' interaction patterns (Figure~\ref{fig:counsellor-temporal}), the most dominant pattern throughout the conversation is \textbf{Analysis / Clarification - Reflection (factual)}, which is particularly prevalent at the beginning of the conversation in both Human-Human and Human-LLM conditions. However, in Human-LLM interactions this pattern peaks early (around 10-20\%) and then gradually decreases, whereas in Human-Human interactions it remains more stable throughout the session. This is consistent with the results of the general pattern distribution, where factual reflection was more frequent in Human-LLM interactions, probably because LLM-generated clients disclose their problems more directly. \textbf{Help, Problem Solving} and \textbf{Moderation} increase as the conversation progresses, especially in Human-Human interactions. This suggests that in the latter stages, human counsellors focus on leading the discussion and addressing the client's problems. In contrast, these patterns are less pronounced in Human-LLM interactions, suggesting a different conversation dynamic where problem solving does not escalate as much over time.
Another noticeable trend is in category \textbf{Generating Motivation}, which remains relatively low throughout the interview but increases slightly towards the end. This could indicate that counsellors are making an effort to introduce motivational interventions before the end of the counselling.

In the client interaction patterns (Figure \ref{fig:client-temporal}), \textbf{Analysis and Clarification of Problems} is the predominant category, especially at the beginning of the conversation. This trend is much more pronounced in Human-Human interactions, where problem elaboration starts high (over 8\%) and gradually decreases over time. In contrast, LLM-simulated clients disclose their problems more immediately but with less intensity, which is consistent with the results of the overall distribution that showed a lower percentage for this pattern in LLM-generated conversations. A key difference between human and LLM clients is the increase in helping, problem solving, and resource activation in the second half of the conversation, especially in the Human-LLM interactions. This suggests that LLM-generated clients are more inclined to shift to problem-solving and resource-orientated discussions as the conversation progresses, whereas human clients may continue to reflect on their problems. In addition, formality at the beginning is higher  of Human-LLM interactions but quickly decreases, while formality at the end increases sharply in the last decile. This reflects the structured nature of LLM-generated responses, which tend to adhere to clear opening and closing conventions.

\subsection{Semantic Analysis}

To further examine differences in conversational patterns, we applied embedding-based representations using the GBERT-Large Client and Counsellor Classifiers. These models, initially trained for sentence classification, were also used to generate embeddings for client and counsellor messages. The resulting embeddings were visualised using t-SNE to illustrate the clustering of different interaction patterns.

The t-SNE projection for client messages (Figure 4) shows that \textbf{Analysis \& Clarification of Problems} (orange) is the most dispersed category, indicating that problem elaboration occurs across diverse conversational contexts. In addition to the static t-SNE plot analysis an interactive version has been analysed. It reveals that the orange cluster at the bottom of the client projection corresponds to \textbf{General Inquiries}, which appears more frequently in Human-Human interactions. This suggests that human clients tend to ask general questions more often than LLM clients, possibly because LLM-generated responses are shaped by instruction-following rather than reciprocal questioning. Additionally, \textbf{Agreement} statements appear as a distinct cluster, with LLM-generated agreements tending to be embedded in longer sentences rather than short affirmations. This aligns with our previous findings that human clients engage more extensively in problem clarification, while LLM-simulated clients disclose problems more directly. Another visible cluster in the center of the projection represents negative feedback on prior problem-solving efforts, further suggesting that LLM clients engage in a more structured feedback process compared to human clients. \textbf{Formalities at the Beginning} form a distinct, tightly clustered region, suggesting that these statements play a structurally separate role from the core counselling discussion. \textbf{Formalities at the End} exhibit a similar concentrated pattern, reinforcing the notion that LLM-generated clients tend to follow clear opening and closing conventions, as observed in the pattern distribution.

\begin{figure}
    \centering
    \includegraphics[width=1\linewidth]{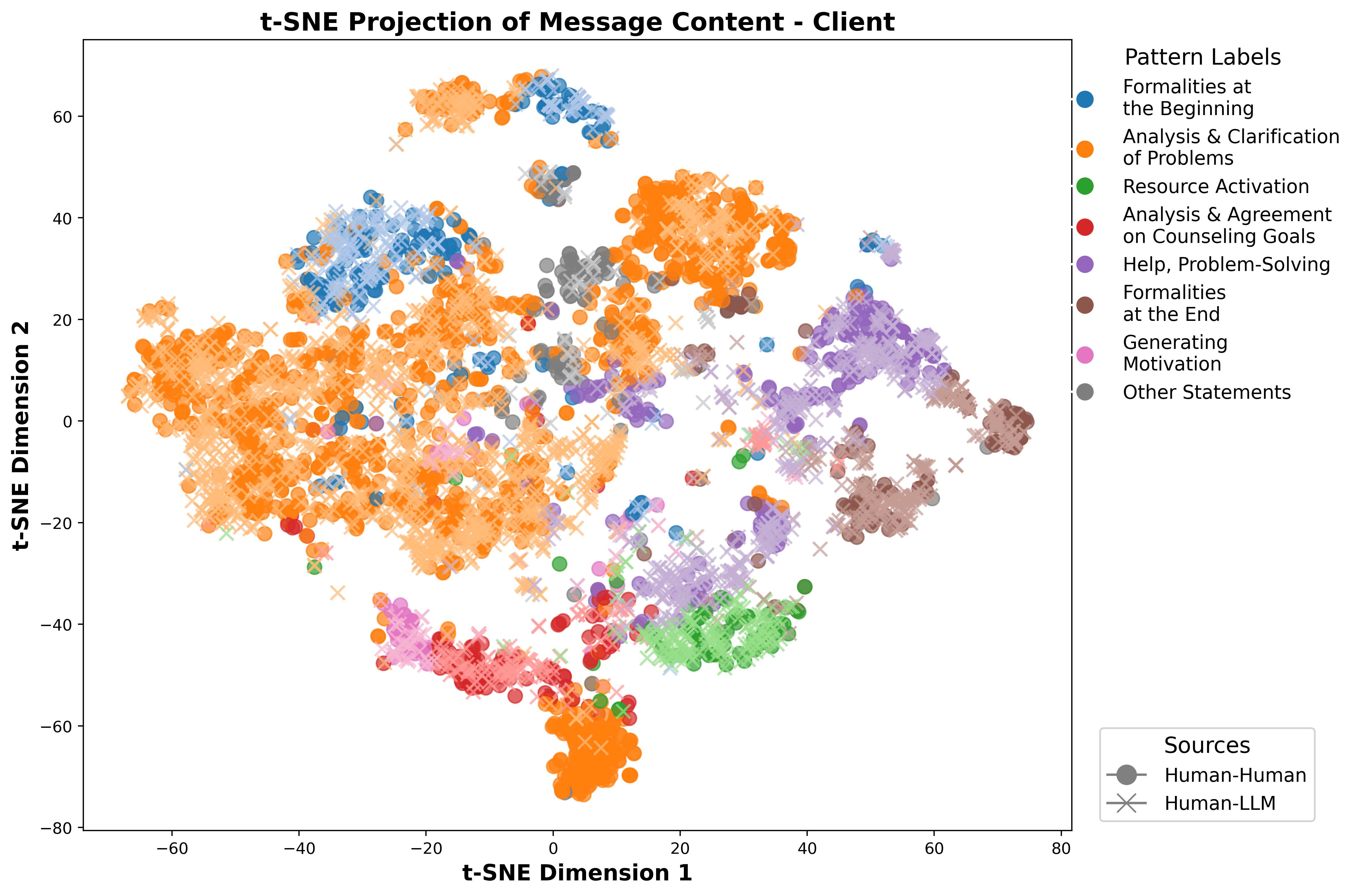}
    \label{fig:tsne-client}
    \caption{t-SNE Projection of Client Sentences }
    \Description{A t-SNE scatter plot of client message embeddings, colored by GeCCo category. The Analysis & Clarification of Problems points (orange) are dispersed across the plot, indicating diverse contexts. Tight clusters appear for Formalities at Beginning (top right) and Formalities at End (bottom left). Distinct clusters for Agreement, Help/Problem-Solving, Resource Activation, and Generating Motivation are also visible, demonstrating semantic grouping of client utterances.}
\end{figure}

\begin{figure}
    \centering
    \includegraphics[width=1\linewidth]{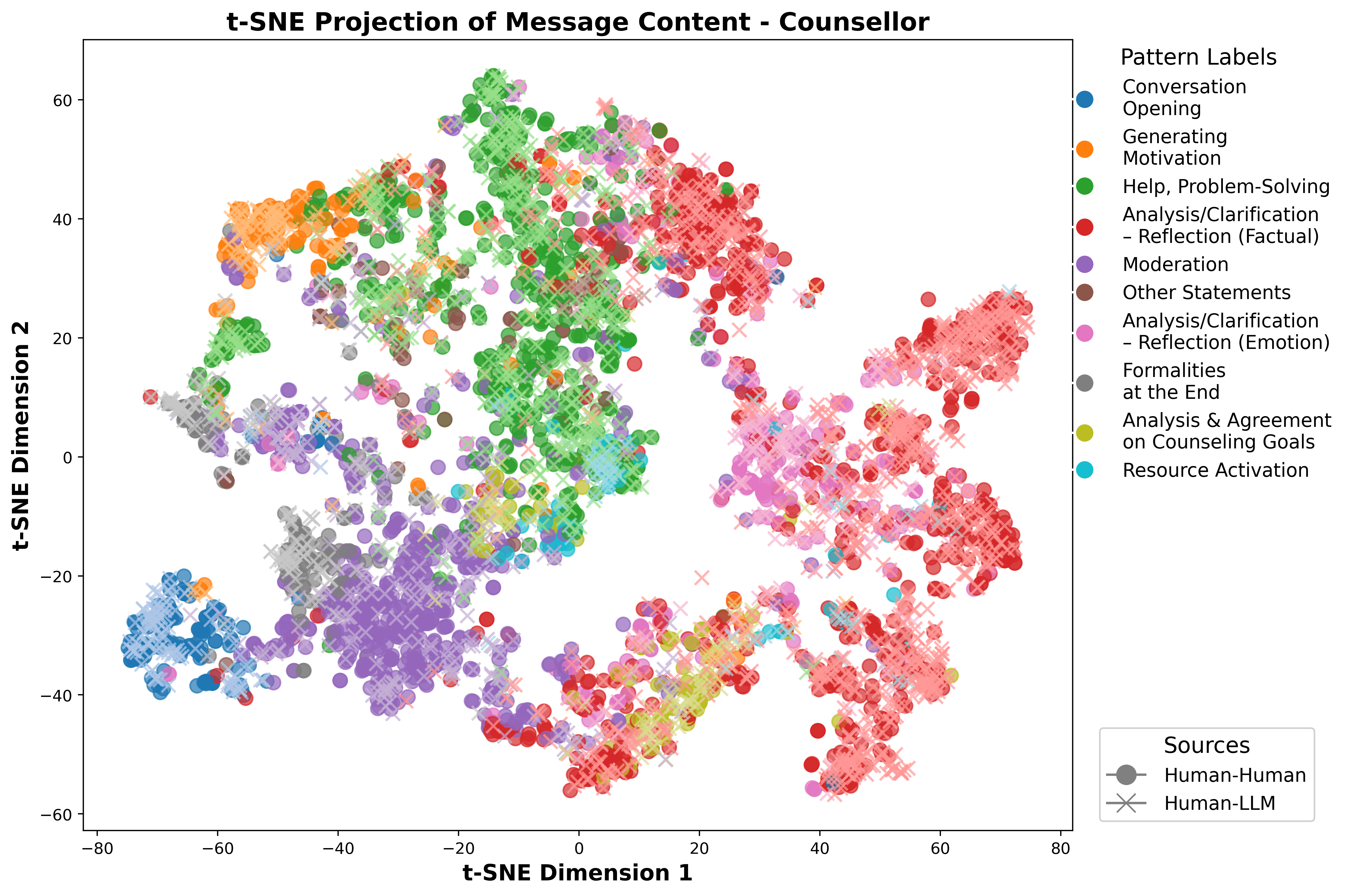}
    \caption{t-SNE Projection of Counsellor Sentences}
    \label{fig:tsne-counsellor}
    \Description{A t-SNE scatter plot of counsellor message embeddings, colored by GeCCo category. Clusters for Help/Problem-Solving are clearly separated, while Reflection (Factual) and Reflection (Emotion) clusters lie adjacent, showing overlap in reflective responses. A separate cluster for Moderation appears, and smaller tight clusters for Conversation Opening and Formalities at Beginning/End indicate structurally distinct utterances.}
\end{figure}

The t-SNE projection for counsellor messages (Figure \ref{fig:tsne-counsellor}) reveals well-defined clusters for \textbf{Help, Problem-Solving}, reflecting the central role of problem resolution in counselling interactions. \textbf{Analysis/Clarification - Reflection (Factual)} and \textbf{Analysis/Clarification - Reflection (Emotion)} are positioned closely, indicating that factual and emotional reflections frequently co-occur within conversations. A key distinction in Human-LLM interactions emerges in the \textbf{Analysis/Clarification (Factual)} category. An interactive plot analysis reveals that when engaging with human clients, counsellors tend to provide more detailed follow-up questions regarding past solution attempts. Examples include inquiries such as "How did he join your circle of friends?" or "What was it like when you weren’t arguing so often?" This pattern is evident in the lower-right region of the t-SNE plot. Another significant finding is the clustering of \textbf{Moderation} statements, which appear more frequently in Human-Human conversations. These messages include procedural clarifications such as "The chat has a time frame of 60 minutes" or "Let’s take turns writing so there is no confusion." This suggests that Human-Human counselling requires more explicit structuring, possibly because human clients exhibit less adherence to a predefined conversational flow compared to LLM-simulated clients. Additionally, the use of emojis was not possible on the learning platform, which may have influenced the nature of moderation-related messages.

\subsection{Educational Implications}
Learning counselling with an LLM has its upsides and downsides. On the upside, LLM-simulated clients offer a highly structured and consistent environment that can streamline training. Their immediate disclosure of issues and solution-focused responses allow trainees to quickly engage in problem-solving and to improve their factual reflection skills. This consistency can be especially beneficial in early training stages or in scenarios where repeated practice is essential for skill development. On the downside, the structured nature of LLM-generated responses may limit exposure to the rich, spontaneous, and emotionally nuanced interactions that characterise human counselling. The reduced emphasis on elaborate problem clarification and emotional attunement may not fully prepare trainees for the unpredictability of real-world counselling sessions. Consequently, while LLM-based role-plays serve as a valuable supplement by enhancing efficiency and standardization, they also risk oversimplifying the complex interpersonal dynamics necessary for comprehensive counselling training. In summary, integrating LLMs into counselling education presents a promising opportunity to expand training accessibility and consistency, but it should be balanced with traditional human interactions to ensure the development of both technical and interpersonal competencies.

\section{Limitations}

While our study offers insightful comparisons between Human-Human and Human-LLM counselling interactions, several limitations should be acknowledged. First, the analysis relied on automated classifiers (GBERT Large) rather than manual annotations, which may not capture every subtle nuance in conversational dynamics. Second, we used role-play scenarios rather than real-world counselling sessions, meaning the findings reflect the dynamics of simulated environments. Third, the analysis focused primarily on the rather coarse categorisation level 3 of the GeCCo system; examining more detailed sub-categories might reveal further nuances. Additionally, it is important to note that our study employed only one large language model -- Mixtral 8x7B with 4-bit quantisation. This choice offers the advantage of better comparability across sessions; however, it also means that alternative LLMs might exhibit different conversational patterns. 

\section{Conclusion and Future Work}
The present study provides valuable insights into the differences and similarities between Human-Human and Human-LLM counselling role-plays. Our results indicate that LLM-simulated clients tend to disclose their issues in a more immediate and structured manner, thereby prompting counsellors to rely more heavily on factual reflection rather than on facilitative moderation. In contrast, human clients typically engage in a more elaborative problem clarification, leading counsellors to adopt a more active moderating role. Furthermore, the temporal and semantic analyses reveal distinct conversational dynamics and clustering patterns between the two interaction types, highlighting the potential and current limitations of using LLMs as simulated clients in counselling training.

For future work, several paths could be pursued to bridge the gap between LLM-simulated and human clients. Refinements in LLM training—such as fine-tuning with more representative conversational data or leveraging reinforcement learning approaches like Direct Preference Optimization (DPO)—could enhance the naturalness and emotional authenticity of simulated interactions. Additionally, incorporating manual annotation and establishing a counselling communications gold standard would provide a more robust framework for evaluation. Working with more detailed categories for utterance classification and exploring diverse linguistic and cultural contexts could further validate and generalise these findings.

\bibliographystyle{abbrv}
\bibliography{references1}
\end{document}